\documentclass[letterpaper,10 pt,conference]{ieeeconf}  

\usepackage{subfig}
\usepackage{graphicx}
\usepackage{comment}
\usepackage{xcolor}
\usepackage{amssymb}

\usepackage[longend]{algorithm2e}
\makeatletter
\renewcommand{\@algocf@capt@plain}{above}
\makeatother
\usepackage{tcolorbox}

\IEEEoverridecommandlockouts                              

\title{\LARGE \bf
Reinforcement Learning-Based Bionic Reflex Control for Anthropomorphic Robotic Grasping exploiting Domain Randomization
}
\author{Hirakjyoti Basumatary
, Daksh Adhar
, Atharva Shrawge
, Prathamesh Kanbaskar
\hspace{0.1em} and Shyamanta M. Hazarika
\thanks{
The authors are with the Biomimetic Robotics and Artificial Intelligence Laboratory (BRAIL) of Department of Mechanical Engineering, Indian Institute of Technology Guwahati, Guwahati 781039, India.
}
\thanks{This work has been submitted to the IEEE for possible publication. Copyright may be transferred without notice, after which this version may no longer be accessible.
}
}
\begin{document}
\maketitle

\thispagestyle{empty}
\pagestyle{empty}

\begin{abstract}

Achieving human-level dexterity in robotic grasping remains a challenging endeavor. Robotic hands frequently encounter slippage and deformation during object manipulation, issues rarely encountered by humans due to their sensory receptors, experiential learning, and motor memory. The emulation of the human grasping reflex within robotic hands is referred to as the ``bionic reflex". Past endeavors in the realm of bionic reflex control predominantly relied on model-based and supervised learning approaches, necessitating human intervention during thresholding and labeling tasks. In this study, we introduce an innovative bionic reflex control pipeline, leveraging reinforcement learning (RL); thereby eliminating the need for human intervention during control design. Our proposed bionic reflex controller has been designed and tested on an anthropomorphic hand, manipulating deformable objects in the PyBullet physics simulator, incorporating domain randomization (DR) for enhanced Sim2Real transferability. Our findings underscore the promise of RL as a potent tool for advancing bionic reflex control within anthropomorphic robotic hands. We anticipate that this autonomous, RL-based bionic reflex controller will catalyze the development of dependable and highly efficient robotic and prosthetic hands, revolutionizing human-robot interaction and assistive technologies.

\end{abstract}

\section{INTRODUCTION}


The human hand, nature's masterpiece for interacting with the world, has inspired robotic grasping in fields like rehabilitation robotics, automation and manufacturing \cite{mattar2013survey}. This technology empowers robots to engage dynamically with the physical world, enabling crucial tasks in assembly lines, handling hazardous materials, and everyday object manipulation. However, the challenges of slippage due to external disturbances and suboptimal grasp forces leading to object deformation are significant obstacles affecting the efficiency and safety of robotic grasping applications \cite{zhu2022challenges}. Addressing the menace of slippage and deformation is paramount to the evolution of robotic grasping. In the realm of prosthetic and robotic hands, this quest leads us to the frontier of bionic reflex, an emulation of the human grasp reflex control.

Model-based strategies and traditional methods have shown limitations in dealing with grasping tasks' complexity. Conventional slippage prevention approaches deal with the friction cone constraints, reliant on precise knowledge of object friction coefficients \cite{romeo2020methods}. Traditional deformation prevention techniques rely on Hooke's law \cite{delgado2017hand} and impedance control \cite{song2019tutorial}, demanding exact stiffness information and an accurate model of the plant/environment dynamics. Computer vision has also been leveraged to detect slips and deformations in grasped objects, but suffer from the problem of occlusions \cite{sui2022incipient}. Even as machine learning has stepped in, its struggle to generalize across diverse shapes and materials persists; 
mainly because supervised learning techniques mandate thresholding shear force signals or deformation levels, necessitating human intervention to label training data \cite{romeo2021method}. 

Traditional methods fall short in autonomously designing bionic reflex control for grasping. RL offers promise by enabling agents to adapt and learn task execution through interaction with the environment. However, RL agents trained in simulations often struggle to perform effectively in the real world due to the ``Sim2Real" gap  \cite{salvato2021crossing}, marked by substantial differences in physics, sensory input, dynamics, and environmental conditions. To address this challenge, DR introduces variability in simulations, making them more representative of real-world complexities \cite{muratore2022robot}. In this study, we present a novel approach for preventing deformation during object grasping, featuring three key contributions: (1) a novel control pipeline utilizing RL for bionic reflex control, (2) eliminating the need for human supervision in setting slip and deformation thresholds during control design, and (3) leveraging DR to enhance adaptability to object properties and environmental shifts, reducing the Sim2Real gap.

\section{Related Work}

Detecting slip is one of the crucial tasks to prevent slippage and attain bionic reflex control. Methodologies of slip detection are mainly based on: vibration, friction model, contact relationship, vision, and data-driven approaches. Vibration-based methods use sensors like polyvinylidene-fluoride (PVDF) film or accelerometers to detect induced vibrations during slippage. Friction model-based approaches rely on maintaining a low tangential force-to-normal force ratio ($<\mu_s$). Contact relationship-based methods involve time-frequency analysis (e.g., STFT, DWT) to detect sudden changes. Vision-based techniques employ optical and tactile sensors for detecting displacement, force distribution, and slippage velocities. Data-driven methods leverage supervised learning methods and deep learning for slip detection \cite{romeo2020methods}.

For the second bionic reflex feature, i.e., deformation prevention, its control necessitates stiffness detection or deformation measurement. Control design of this bionic reflex feature is quite challenging and is still an open research problem \cite{zhu2022challenges}. Existing methodologies for stiffness detection and control in the literature include: (1) intrinsic vibration frequency-based signal processing, (2) time-domain analysis methods (3) the integration of measuring devices \cite{zhang2022stiffness} (4) Hooke's Law \cite{andrecioli2013adaptive} and (5) Impedance Control \cite{ji2021grasping}. In \cite{deng2016slippage}, stiffness was identified and controlled using a PVDF sensor, but it relied on human supervision to establish voltage thresholds for categorizing deformable objects. Slippage detection was based on the empirical mode decomposition (EMD) of the force sensor signal. Deformation control utilizing Hooke's law and impedance based control requires exact stiffness knowledge and desired model references respectively for their effective implementation.

Some studies employ vision-based techniques for deformation detection \cite{zhu2022challenges, cretu2011soft, makihara2022grasp}. Tactile sensors like the Gelsight sensor offer high-resolution tactile data by measuring elastomer deformation \cite{zhu2022challenges}. However, such sensors can be costly and less accessible. \cite{makihara2022grasp} relies on image processing techniques like pixel analysis to generate a stiffness map using the `pix2stiffness' method. This map guides grasp pose detection for preventing damage to unknown deformable objects. However, stiffness map generation is not automated, and no force control based on contact dynamics to minimize deformation was considered.

Some non-conventional approaches to tackle deformation control include relying on kinematics to detect stiffness in underactuated mechanisms \cite{zhang2022stiffness}; while others use tactile sensors to determine slippage and control deformation by manipulating object weight through reorientation \cite{kaboli2016tactile}. Bistable compliant underactuated grippers enhance deformable object grasping \cite{mouaze2022bistable}. Soft grippers offer adaptability to objects of diverse shapes by leveraging material compliance \cite{wang2017shape}, but come with limitations such as variability, miniaturization potential, high complexity and costs \cite{milojevic2021novel}. Typical rigid robotic hands struggle with deformable objects due to their fixed stiffness. To overcome this limitation and achieve adaptable grasping, we focus on implementing an active control system, which is the primary objective of this article.

\section{Fundamentals}

\subsection{Reinforcement Learning}

The RL problem is usually formulated as a Markov Decision Process (MDP) described by the tuple $(S,A,P,R,\gamma)$, where $S$ is the set of states of the robot and environment at time $t$, ${s_t} \in S \subseteq {R^n}$ and $A$ is the set of actions, ${a_t} \in A \subseteq {R^m}$, that the agent takes according to policy $\pi ({a_t}|{s_t})$ to maximize the reward $R$ = $\sum\limits_{i = t}^T {{\gamma ^{i - t}}r({s_i},{a_i})}$. $P({s_{t + 1}}|{s_t},{a_t})$ is the state transition function and $\gamma  \in [0,1]$ is the discount factor which determines the importance of short-term rewards. The optimal policy which maximizes the expected cumulative return is given as:  
\begin{equation}
    \label{reinforcement_learning}
    {\pi ^*} = \mathop {\arg \max }\limits_\pi  \sum\limits_{t = 0}^T {{E_{({s_t},{a_t}) \sim {\rho _\pi }}}\left[ {{\gamma ^t}r({s_t},{a_t})} \right]} 
\end{equation}

where $\rho_{\pi}(s_t,a_t)$ is the distribution of the trajectory $(s_1,a_1,...,s_T,a_T)$ produced by the policy, $\pi$.

\subsection{Actor critic model}

One of the RL algorithms is the actor-critic model. Actor-critic is a popular approach in the field of RL because it combines the benefits of both policy-based methods (like the Actor) and value-based methods (like the Critic), addressing some limitations of each approach individually \cite{haarnoja2018soft}. Some of the reasons why Actor-Critic methods are often considered better or more advantageous compared to other RL algorithms are: bias-variance trade-off, sample efficiency, convergence, stability, generalization across continuous action space, exploration-exploitation capability, policy improvement theorem implementation, etc. 

 \begin{figure*}[ht!]
  \includegraphics[width= 1 \linewidth]{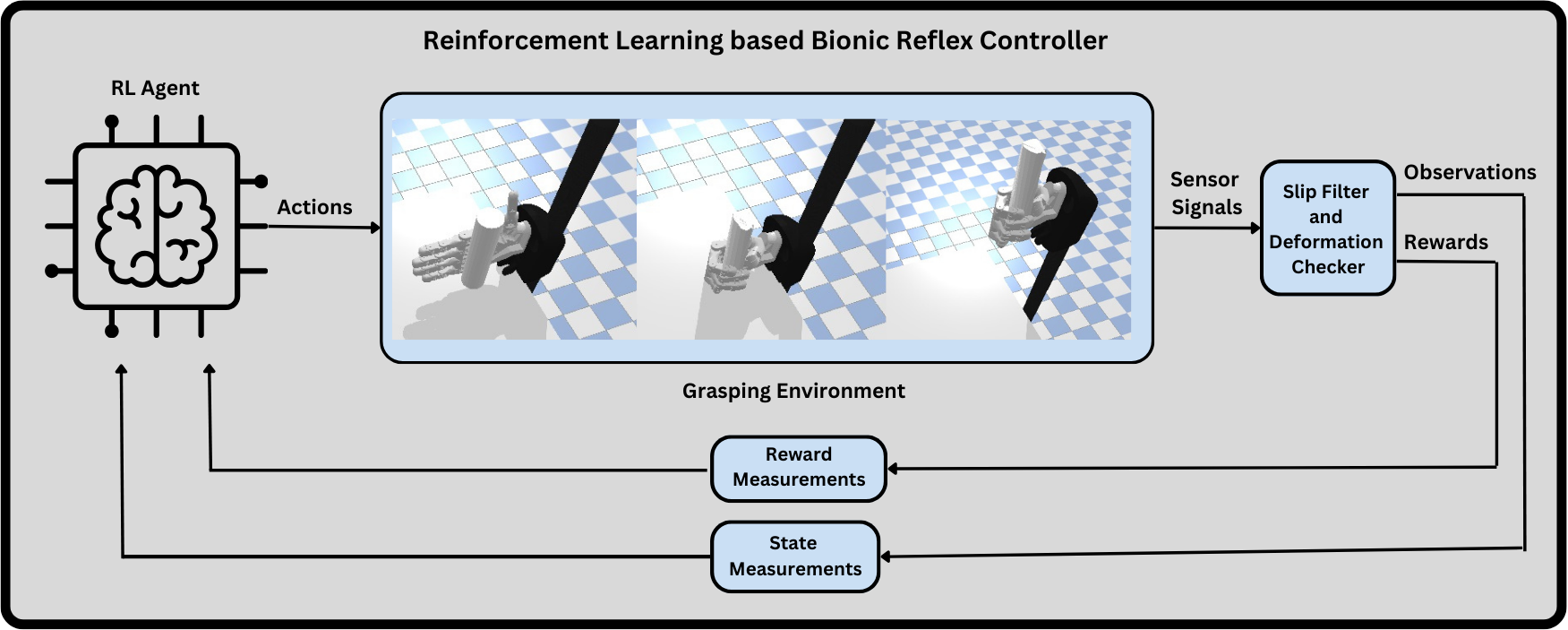}
  \centering
  \caption{RL based Bionic Reflex Control Pipeline}
  \label{RL_bionic_reflex_pipeline}
\end{figure*}

\subsubsection{Soft Actor-Critic}

The Soft Actor-Critic (SAC) algorithm, introduced by \cite{haarnoja2018soft}, is an off-policy actor-critic method. SAC has demonstrated notable effectiveness in tackling continuous control challenges and exhibits robustness when confronted with variations in the environment. This algorithm deals with the maximum entropy  RL problem, an extension of the conventional RL objective (as defined in Equation (\ref{reinforcement_learning})), by incorporating an entropy term.

\vspace{-1em}

\begin{equation}
    {\pi ^*} = \mathop {\arg \max }\limits_\pi  \sum\limits_{t = 0}^T {{E_{({s_t},{a_t}) \sim {\rho _\pi }}}\left[ {{\gamma ^t}r({s_t},{a_t}) + \alpha H(\pi ( \cdot |{s_t}))} \right]} 
\end{equation}

where, $H(\pi(\cdot|s_t))$ is the entropy associated with the action distribution, $\alpha$ is the temperature parameter, regulating the significance of the entropy component. This objective optimizes both the expected return and action entropy simultaneously, fostering exploration and encompassing the acquisition of multiple actions that are close to optimal \cite{haarnoja2018soft}.

The SAC algorithm utilizes an actor-critic policy search approach, comprising a policy $\pi_\phi$ serving as the actor and a soft Q-function, $Q_\theta$ acting as the critic. In the learning process, $Q_\theta$ undergoes iterative updates to approximate the temporal difference (TD) target as follows:

\begin{equation}
    {\hat y_t} = r({s_t},{a_t}) + \gamma {E_{{s_{t + 1}} \sim p}}[{V_\theta }({s_{t + 1}})]
\end{equation}

where ${V_\theta }({s_t}) = {E_{{a_t} \sim \pi }}\left[ {{Q_\theta }({s_t},{a_t}) - \alpha \log (\pi ({a_t}|{s_t}))} \right]$ is the state value function. Hence, $Q_\theta$ can be updated by minimizing the loss:
\begin{equation}
    {J_Q}(\theta ) = {E_{({s_t},{a_t}) \sim D}}\left[ {\frac{1}{2}{{({Q_\theta }({s_t},{a_t}) - {{\hat y}_t})}^2}} \right]
\end{equation}

and the policy $\pi_\phi$ can be updated by minimizing the loss
\begin{equation}
    {J_\pi }(\phi ) = {E_{{s_t} \sim D}}\left[ {{E_{{a_t} \sim {\pi _\phi }}}\left[ { - {Q_\theta }({s_t},{a_t}) + \alpha \log (\pi ({a_t}|{s_t}))} \right]} \right]
\end{equation}

where $D$ is the experience replay buffer.

\subsection{Discrete Wavelet Transform - Haar Wavelet}

Wavelet analysis, using the Discrete Wavelet Transform (DWT), is a standard method for efficiently examining localized power variations in time series data. In a one-dimensional time series f(t), it can be represented as:

\begin{equation}
    f(t;{F_N},\alpha) = \sum\limits_{k \in Z} {{c_{{j_0},k}}{\varphi _{{j_0},k}}(t) + \sum\limits_{k \in Z} {\sum\limits_{j = {j_0}}^\infty  {{d_{j,k}}} } } {\psi _{j,k}}(t)
\end{equation}

where $f(t; F_N, \alpha)$ is influenced by the normal force $F_N$ and the sensor parameter $\alpha$. The other terms involved include $\varphi$ (scaling function), $\psi$ (wavelet function), $k$ (shift parameter), $j$ (scale parameter), $\{{c_{{j_0},k}}\}_{k \in Z}$ (approximate coefficient sequence), and $\{{d_{{j},k}}\}_{k \in Z}$ (detail coefficient sequence). $\sum\limits_{k \in Z}{c_{{j_0},k}}{\varphi _{{j_0},k}}(t)$ provides a low resolution approximation of $f(t;{F_N},\alpha)$, while $\sum\limits_{k \in Z} {\sum\limits_{j = {j_0}}^\infty  {{d_{j,k}}} } {\psi _{j,k}}(t)$ depicts the details of $ f(t;{F_N},\alpha)$ in high resolution \cite{yang2015new}. Since, slip is a transient but rapid change in signal, hence in this paper we focus on high-resolution details of the sensor signals. Next, we need to choose an appropriate wavelet function.


Haar Wavelet is one of the common techniques to detect slippage while grasping an object \cite{zhang2022design}. It is usually chosen to differentiate slippages because slip signals are transient changes in original signals, and Haar wavelet is itself discontinuous, non-differentiable, and asymmetric. Hence, slip signals are well reflected and captured by the high-frequency components of the Haar wavelet \cite{yang2015new}. Haar wavelet transform is used for slip detection in robot grasping because of its ability to analyze changes in signal patterns at different scales. Its advantages include multi-resolution analysis, the ability to detect localized changes, real-time processing, the ability to threshold, simplicity and speed, directional information, sparse representation, robustness to noise, low memory requirements, etc. In the current work, Haar wavelet is hence utilized to detect slippage while grasping an object.

\section{Design Methodology}

\begin{figure*}[ht!]
  \includegraphics[height=0.4\linewidth]{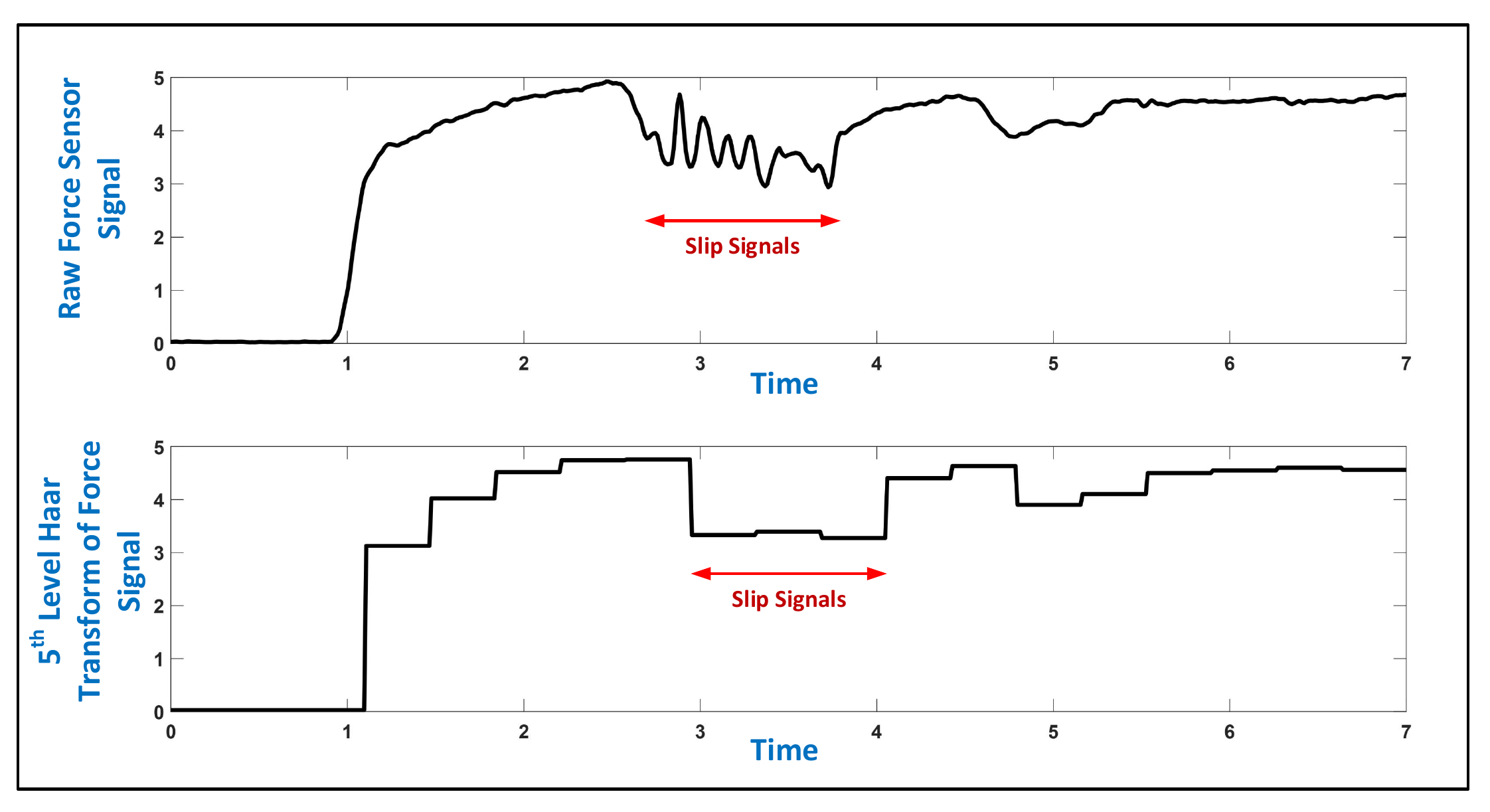}
  \centering
  \caption{Force sensor signal while grasping and lifting an object. Fifth level Haar decomposition of raw force sensor signal is used to detect slip. The positive gradient is a reflection of the load being applied. The opposing variation trend is a representation of slip.}
  \label{DWT_Haar_Wavelet}
\end{figure*}

\subsection{PyBullet and OpenAI Gym}

The current study aims to train an anthropomorphic robotic hand to grasp objects and lift it while minimizing slippage and deformation, leveraging RL in PyBullet \cite{collins2021review} (Fig. \ref{RL_bionic_reflex_pipeline}). The RL agent sends joint torques as actions to the robotic hand grasping environment which helps it to grasp and lift the object. The slip filter and the deformation checker (explained in detail in the next section) are utilized to calculate the rewards and generate optimal actions. The custom unified robot description format (URDF) model of the anthropomorphic hand from \cite{basumatary2022design} is used. Torque control is chosen within the PyBullet simulation since it allows precise control over joint motor by adjusting the voltage induced; a preferred mode for replicating real-world bionic device behavior. A custom OpenAI Gym environment is then developed, leveraging PyBullet's control functions within the class functions. In line with the standard structure of a Gym environment, the defining class consists of five essential functions: initializer, step, reset, render, and close.

In the initializer function, we initiate PyBullet connection via `pybullet.connect,' set gravity to $-9.8m/s^2$, and create observation and action spaces. Our Gym environment's observation space includes joint positions, velocities, finger contact forces, slip and deformation states. The action space consists of torque values for the 15 hand joints. The `step' function executes commands based on the simulation's running time. When the environment is active for less than 5 seconds, the RL agent executes the action of grasping. Beyond 5 seconds, it combines grasping and lifting, capping each episode at a maximum of 10 seconds. The `step' function primarily executes actions in PyBullet, retrieves current rewards, and returns the environment's state. This setup ensures smooth interaction between the RL agent and the simulation, enabling the agent to adapt its actions based on the environment state and enhancing grasping and lifting capabilities. The render and close functions are straightforward as the former sets the PyBullet connect mode to GUI, and the latter simply disconnects it. Once the environment class is defined, we can save it as a Gym environment and call it while training and testing our RL agent.

\subsection{Slip Detection }

Slip detection between the object and fingers relies on continuous analysis of fingertip force signals. We create a Haar wavelet of this signal and identify slip (negative gradient) and no-slip (positive gradient) events (see Fig. \ref{DWT_Haar_Wavelet}). In the simulation, we use the PyWavelets library. Initially, we generate a force log with the latest force values at each fingertip. We apply the `wavedec' function with parameters: force\_log, decomposition type (Haar), and decomposition level (5 in our case). The resulting coefficients include an approximation, representing the least varying or most approximated Haar transform used for inverse wavelet reconstruction. Fig. \ref{DWT_Haar_Wavelet} illustrates a reconstructed force signal at the 5th level, depicting a deliberate slipping episode with a notable signal drop. To detect the slippage, we observe the gradient of the Haar wavelet. If this gradient is negative, then there is a slip occurring, and vice versa.

\subsection{Deformation Calculation}
\begin{figure}
  \centering
  \includegraphics[height=0.8\linewidth]{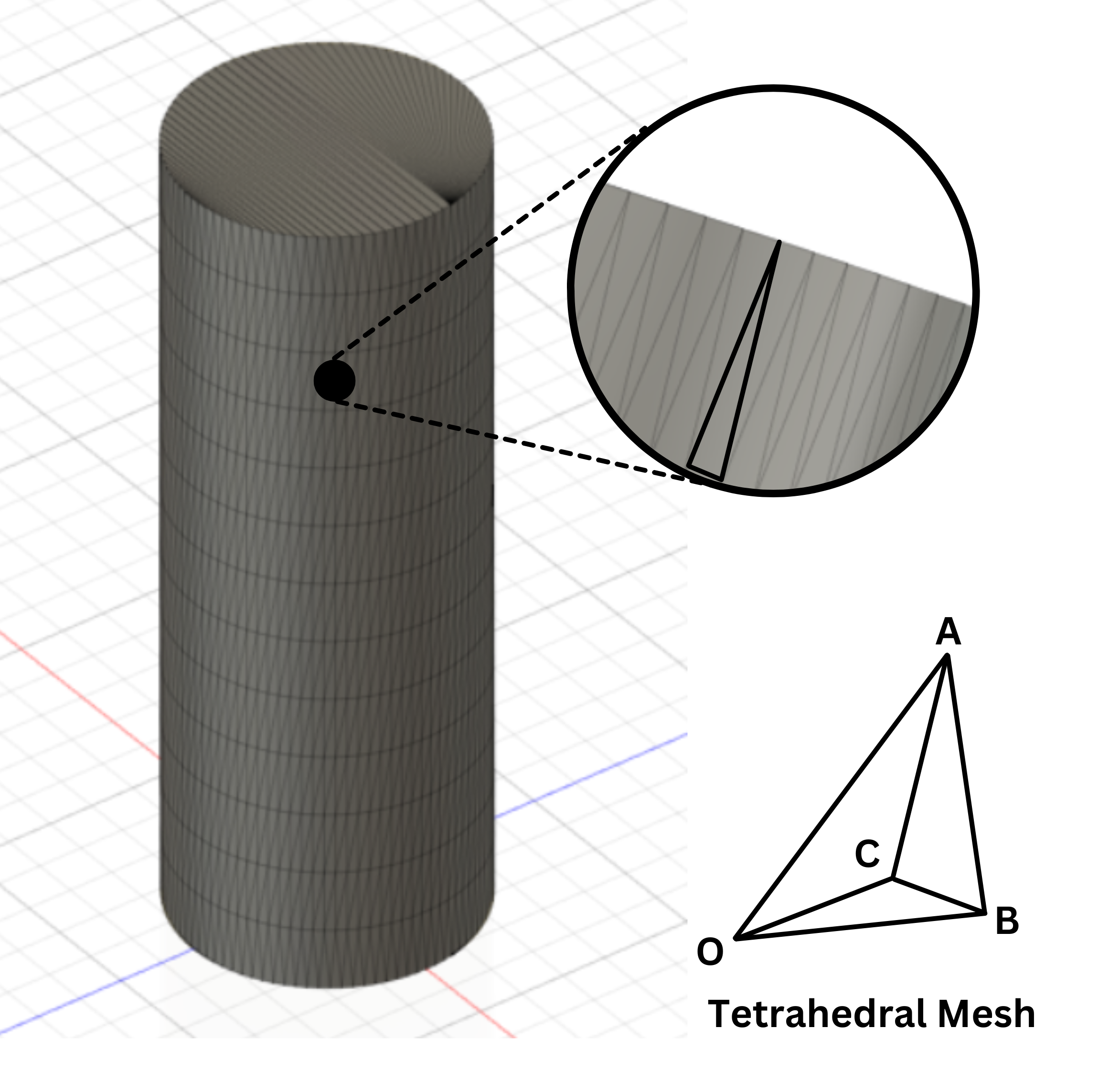}
  \caption{Triangular mesh description of the deformable cylinder. Point \textbf{O} represents reference origin.}
  \label{Deformation_calculation}   
  \vspace{-2em}
\end{figure}

PyBullet represents surface mesh data using vertex and triangle lists (as shown in Fig.\ref{Deformation_calculation}). The PyBullet function ``getMeshData" returns the mesh information (vertice indices) of the triangular meshes that form the 3D object. The reference frame is also changed from the ground to the grasped object (shifted origin) dynamically during grasping and lifting. This step enables us to account for any translations or rotations experienced by the object during manipulation. To estimate volume changes for deformable objects without direct mathematical formulas, we define a general point (shifted origin) to create a tetrahedron (Fig. \ref{Deformation_calculation}) and calculate the signed volume using the following formula:
\begin{equation}
    Signed \hspace{0.5em} Volume = \textbf{AB}. (\textbf{AC} \times \textbf{AO}) / 6
\end{equation}
where points A, B, and C are concurrently selected from the mesh, O is a reference point taken arbitrarily. 
The ``signed volume" indicates the orientation or direction in which the volume is calculated; here, the surface normal of \textbf{ABC} determines the sign and weight of each tetrahedron, and the summation of all such solids captures the object's overall volume. This volume is then used to calculate deformation (change in volume) at every time step.  Our approach provides a dynamic and comprehensive assessment of deformation and allows real-time monitoring for reward calculation.

\subsection{Rewards}
To achieve our goal of a secure grasp with minimal object
deformation, we define the rewards as follows:

\begin{equation}
    \sum\limits_{i = 1}^5 {\left( {\frac{1}{{\ln ({x_i} + 1.1051)}}} \right)}  + \sum\limits_{i = 1}^5 {\left( {{\delta _i}.10} \right)}  - \sum\limits_{i = 1}^5 {\left( {{\theta _i}.10} \right)}  - C \times \Delta 
\end{equation}

The goal here is to slowly but accurately make the hand grasp the object so that it does not slip during lifting while also ensuring that this grasp does not damage the object through deformation. The first term uses an inverse logarithmic relation to define the reward for each step. $x_i$ is the distance between the fingertip and the object; as the finger approaches the object, $x_i$ approaches zero and the reward term is maximized. The following term further signifies firm contact. The distance term can give similar rewards even when $x_i$ is close to 0. To ensure that our agent can differentiate between close contact and contact, this discrete term, $\delta$ further increments a +10 reward on contact of each finger. $\delta$ is 1 if there is contact and 0 otherwise. If all goes well, by now, our agent should be able to entrap the object between the bionic fingers and ensure contact with the object. Our next step involves lifting this object with minimal slipping.  Keeping the scaling similar to other parts of the reward, a value of 10 is penalized from the total reward for each finger whenever slipping is detected. The slip value is formulated as a boolean ($\theta$), taking values 1 if slip is detected and 0 otherwise. Although this seems accurate enough, a drawback of our previous three terms is the promotion of firm grasp, i.e., the grasp is tight enough to damage the object. To counter this, a penalty for deformation is also added. To calculate this, we quantify the deformation value from the volume gradient($\Delta$), that is, change in volume, and scale it accordingly ($C = 50/(initial \hspace{0.5em} volume)$). This value is subtracted from our reward function directly, penalizing the agent whenever the grasp causes excessive deformation. The pseudo-code of the algorithm of the RL-based bionic reflex controller is shown in Algorithm \ref{Algo_1}.

\begin{algorithm}
    \caption{Bionic Reflex Control}
    \label{Algo_1}
    \footnotesize
    \textbf{Input:} Grasping environment in PyBullet \\
    \textbf{Output:} Optimal joint torques ($\tau$) to lift the grasped object without slippage and deformation \\
    {
        \ForEach{episode}{
            Initialize $S$ (observations of states from the grasping environment) \;
            \ForEach{step of episode}{
                Generate actions based on \textbf{Soft Actor-Critic} algorithm \;
                Grasp the object \;
                Lift the object \;
                Slip discrimination during object lifting \;
                \eIf{Object Drops}{
                    episode ends \;
                    reset simulation \;
                }{
                    \If{Slip Occurs}{
                        $\tau \gets \tau + \Delta \tau$ \;
                        \eIf{Deformation Occurs}{
                            $\tau \gets \tau + \Delta \tau - \lambda$ \tcp*{(-$\lambda$ is a small decrement in force)}
                        }{
                            $\tau \gets \tau + \Delta \tau + \lambda$ \;
                        }
                        continue \;
                    }
                }
                End lifting motion (Grasp Successful) \;
                \If{Grasp Successful}{
                episode ends \;
                reset simulation \;
            }
            }
            Until Episode Time is terminal  
        }
    }
\end{algorithm}

\begin{figure}[h!]
  \includegraphics[width = 0.9\linewidth]{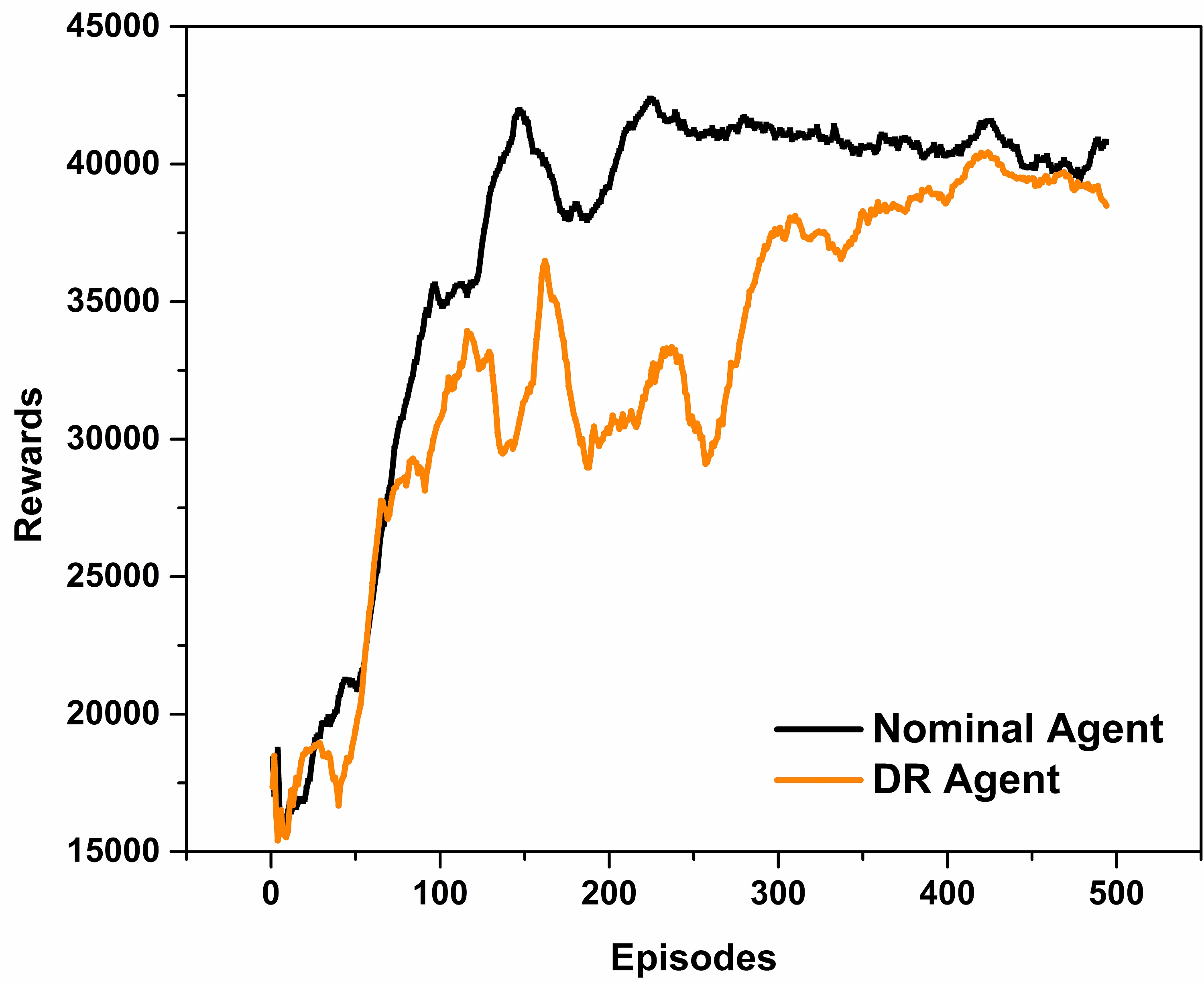}
  \centering
  \caption{Average Reward Plots of Nominal and DR Agent}
  \label{reward_plots}
\end{figure}

\begin{figure*}[h!]%
\centering
\subfloat[][]{%
\label{object_grasp_pose_successful}%
\includegraphics[height=4cm]{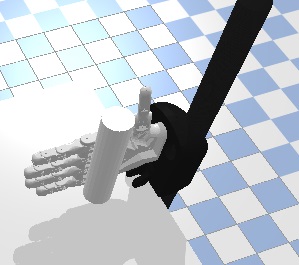}}%
\quad
\subfloat[][]{%
\label{object_grasping_successful}%
\includegraphics[height=4cm]{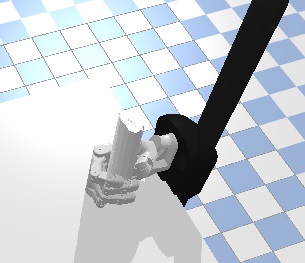}}%
\quad
\subfloat[][]{%
\label{object_lifting_successful}%
\includegraphics[height=4cm]{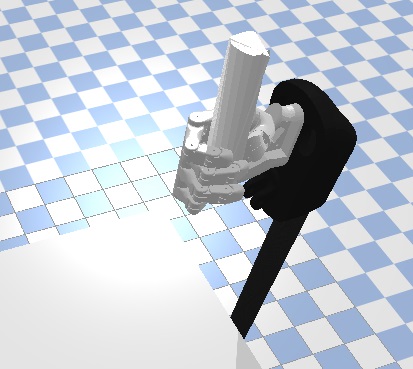}}%
\caption{Learned Grasp simulation in PyBullet: (a) Initial Grasp Pose, (b) Grasping the object (c) Object Lift without Slippage }
\label{learned_grasp_sucessful}
\end{figure*}

\begin{figure*}[h!]
     \centering
     \subfloat[][]{\includegraphics[width=8cm]{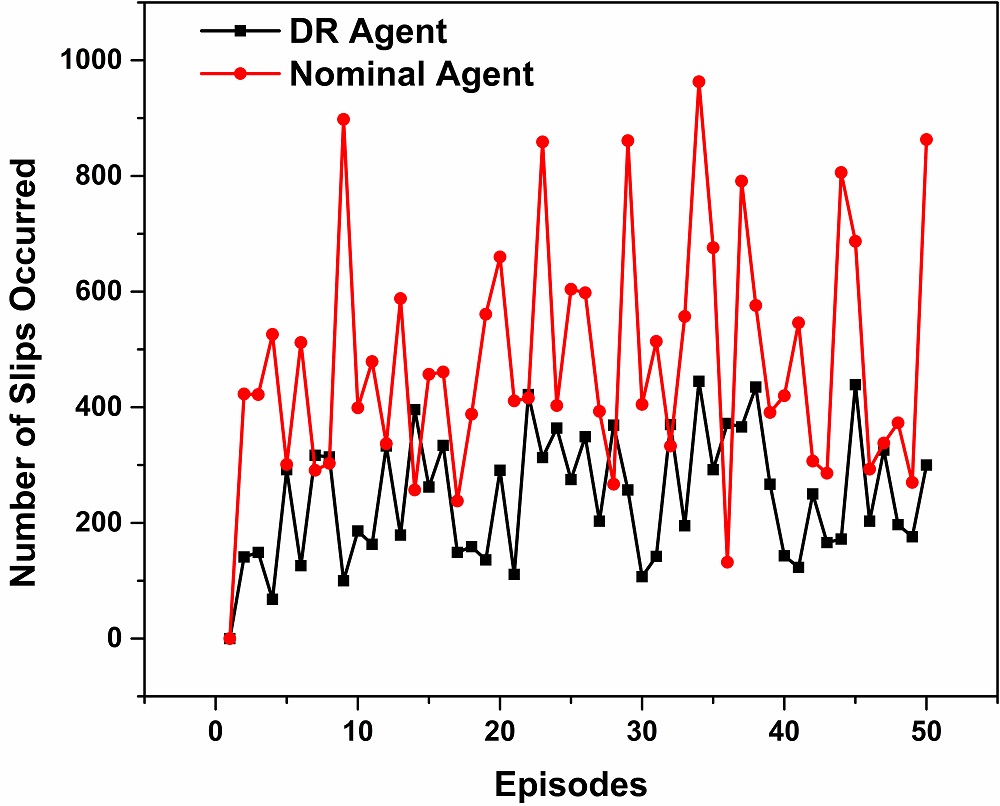}\label{slip_test}}
     \subfloat[][]{\includegraphics[width=8cm]{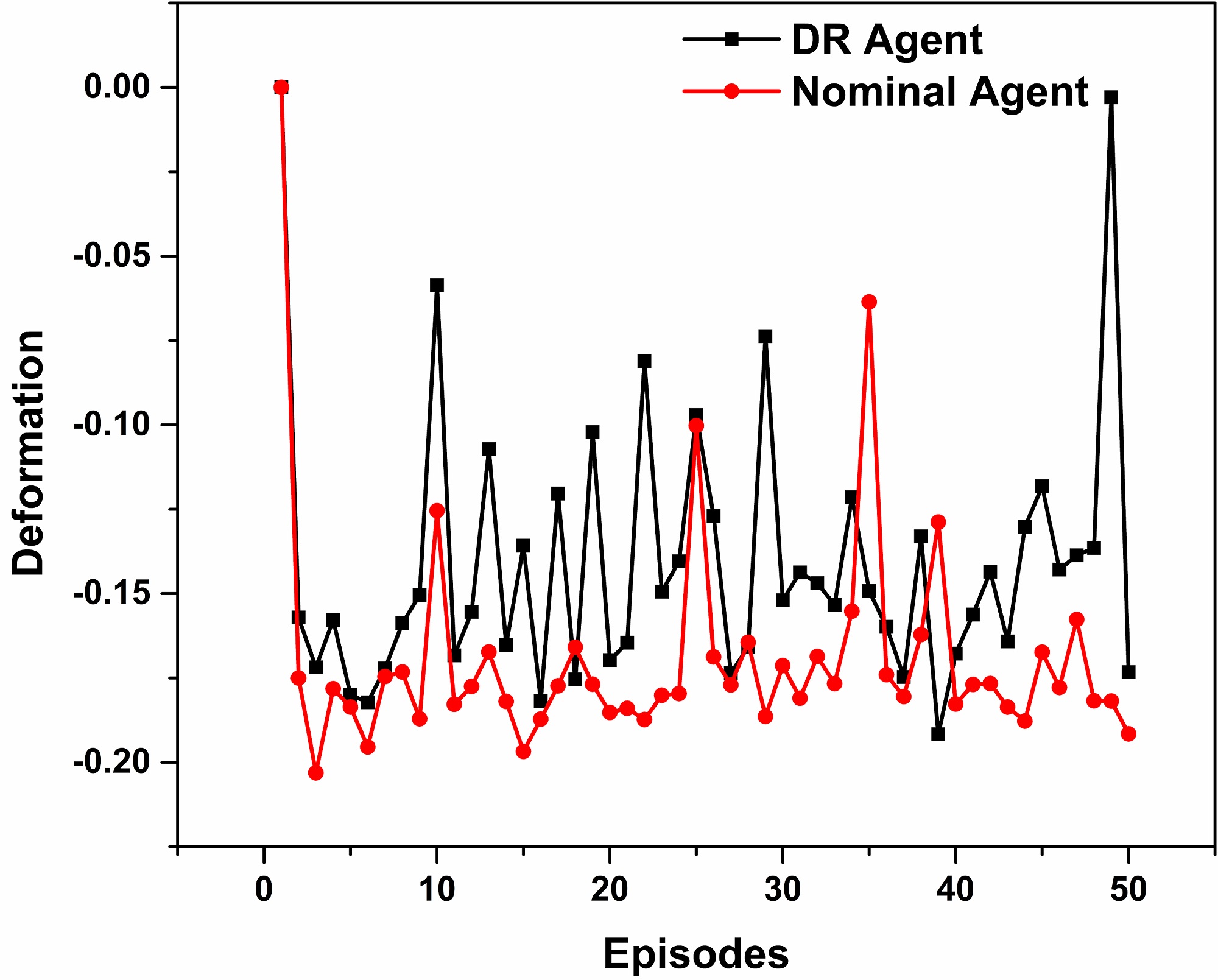}\label{deformation_test}}
     \caption{Performance Tests of success rates on Unseen Objects. Parameters randomized were the object weights, stiffness, and friction coefficients while grasping: (a) Slips prevented by Nominal and DR Agent at unseen objects task. (b) Amount of deformation (in mm) prevented by Nominal and DR Agent on unknown objects task}
     \label{validation_unseen_objects}
\end{figure*}

\section{Results and Discussion}

\subsection{Reward Plots for Nominal and DR Agents}

The reward plots, i.e., the average reward for the RL agents trained in different scenarios (one nominal and other DR), are plotted (Fig. \ref{reward_plots}). The black-colored graph is the reward plot of the RL agent trained in a nominal environment, and the orange-colored graph is the reward plot of the RL agent where the mass, friction coefficient and object stiffness are randomized. The convergence times of the reward plots are nearly similar. However, the rewards of the DR agent are lower than the agents trained in a nominal environment, as there are disturbances due to the randomized parameters, i.e., weight, friction, and stiffness, which creates more slippages and deformations; hence the cumulative reward falls at the beginning. However, the domain-randomized RL agent performs better at untrained and unseen object tasks (discussed in the next section). The trained agent's grasp simulation is shown in Fig. \ref{learned_grasp_sucessful}, and we can see that the agent has been able to grasp and lift the object successfully without any slippage with minimum deformation (Fig. \ref{object_lifting_successful}).

\subsection{Success Rates for Nominal and DR Agent}

Fig. \ref{validation_unseen_objects} shows performance tests of both the agents trained on the nominal environment and as well as on the DR environment on unseen object grasping task. The unknown objects are randomized to have different weights, stiffness, and friction properties.  Fig. \ref{slip_test} shows the success rates of preventing slippage while grasping unknown objects for nominal and domain-randomized agents. The x-axis shows the number of episodes for which the learned agent was tested. The y-axis represents the frequency of slips prevented.  From the success rates plot, it is evident that the agent trained in a domain-randomized environment has been able to prevent more slips than the nominal agent. Fig. \ref{deformation_test} shows the deformation prevention efficiency of the nominal and the DR-trained agent on unknown objects tasks with different object properties. It is seen that while grasping objects with unknown properties, the DR agent is able to grasp the object with lesser deformation than the nominal agent, together with preventing slip and droppage. From both performance tests, we can conclude that the learned DR agent is better equipped to reduce the sim-to-real gap than the nominal agent.

\section{Validation by Grasp Quality Metric}
\begin{figure}[h!]
  \includegraphics[width = 1\linewidth, trim=0 4.5cm 0 2cm, clip]{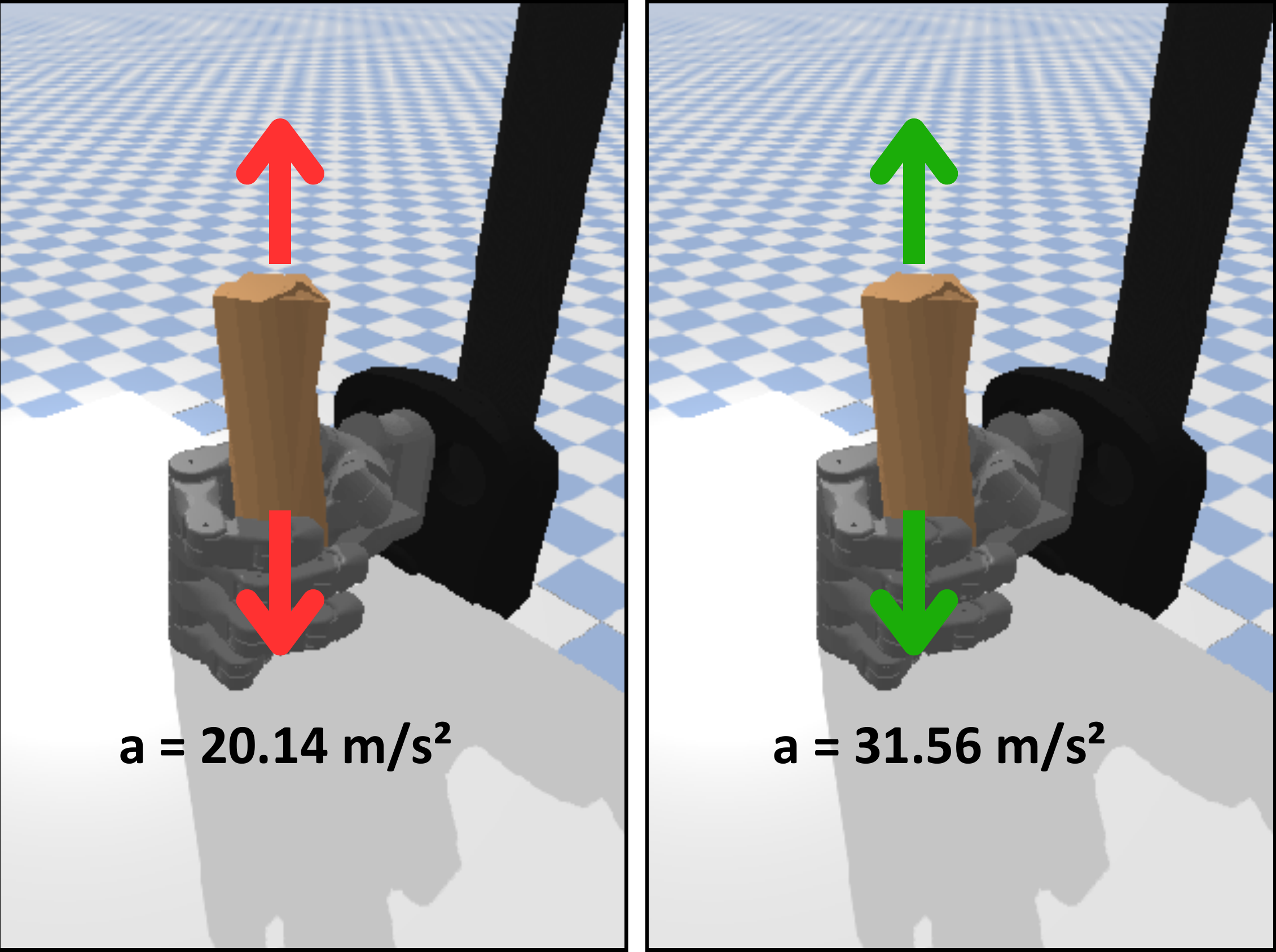}
  \centering
  \caption{Linear instability metric calculation from shake task \cite{huang2022defgraspsim}. The average acceleration before loss of contact for the nominal agent is 20.14 $m/s^2$ (left) and 31.56 $m/s^2$ for the DR agent (right).}
  \label{shake_test}
  \vspace{-2em}
\end{figure}

\begin{figure}[h!]
  \centering
  \includegraphics[height=0.8\linewidth]{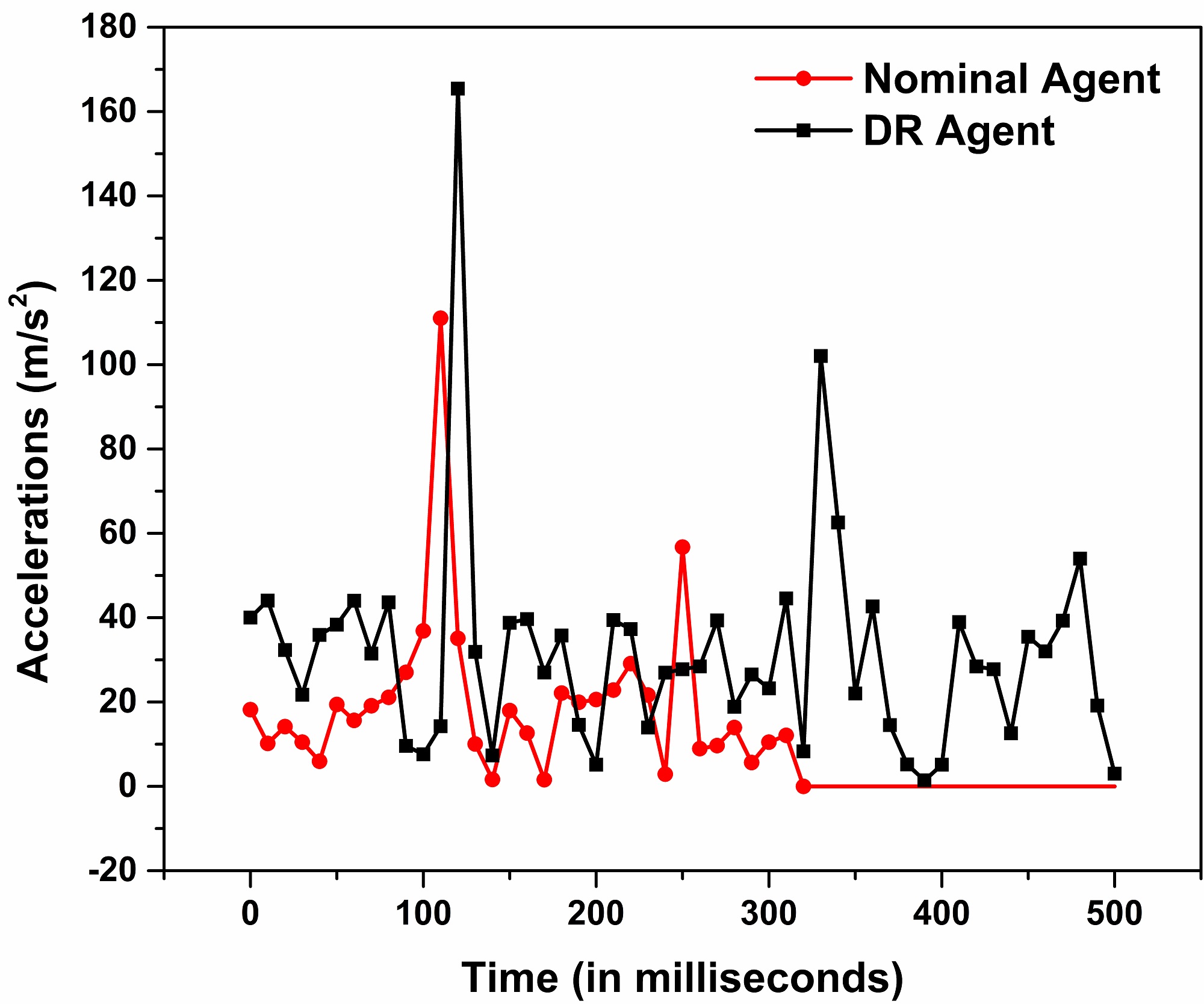}
  \caption{Accelerations of the grasped object during the shake task for the Nominal and DR agent}
  \label{Acceleration_shake_task}   
  \vspace{-2em}
\end{figure}


To assess the grasp quality of deformable objects, we employed the linear instability metric derived from the shake task. This choice was motivated by prior evidence, demonstrating its efficacy in quantifying the effectiveness of grasps on deformable objects \cite{huang2022defgraspsim}. In the context of the shake task, we subjected the grasped object to linear acceleration until it dropped, at which point we computed the linear instability metric by averaging the loss-of-contact accelerations. A higher value of this metric indicates a better grasp, signifying the ability to endure more significant accelerations and maintain efficient object manipulation during abrupt jerks and external disturbances. In our experiments, we incrementally increased the acceleration applied to the grasp while elevating the object until the anthropomorphic robotic hand lost contact with the grasped object. Subsequently, we quantified the acceleration and calculated the linear instability grasp quality metric. Fig. \ref{shake_test} illustrates the outcome of the shake test, revealing that the average acceleration before loss of contact for the nominal RL agent was 20.14 $m/s^2$, while that of the DR agent was 31.56 $m/s^2$. The plot of the accelerations is shown in Fig. \ref{Acceleration_shake_task}. These results imply that the object was more susceptible to displacement under external forces and disturbances in the case of the nominal agent, in contrast to the DR agent, suggesting a higher bionic reflex capacity for the latter.

\section{CONCLUSIONS}

Enabling bionic reflex control for robotic hands is essential in the pursuit of optimizing the efficiency of grasping. This study introduces a novel approach for enhancing the grasping and manipulation capabilities of robotic hands by employing a bionic reflex controller developed using deep RL (DRL). The empirical findings, coupled with the success rates assessment and validation via the shake task, demonstrate the efficacy of the DRL controller in preventing slippage and deformation. Notably, the DRL controller exhibits the capacity to autonomously acquire these preventive skills through interaction, devoid of any prior human-derived knowledge or supervision. This approach holds substantial promise for the advancement of prosthetic hands and robotic grasping research, offering a pathway to design robust, autonomous bionic reflex controllers that reduce the need for human intervention. Future research endeavors will include designing better robust adaptive bionic reflex controllers to facilitate grasping objects with varying properties while also considering the incorporation of diverse randomized parameters.

\addtolength{\textheight}{-12cm}   







\bibliographystyle{IEEEtran}
\bibliography{mybib}

\end{document}